\documentclass[letterpaper, 10 pt, conference, twocolumns]{ieeeconf}

\IEEEoverridecommandlockouts
\usepackage{graphicx}
\usepackage{epsfig}
\usepackage{amsmath}
\usepackage{amssymb}
\usepackage{bbm}
\usepackage{eufrak}
\usepackage{subfig}

\usepackage[noabbrev]{cleveref}

\makeatletter
\def\BState{\State\hskip-\ALG@thistlm}
\makeatother

\DeclareMathOperator{\Ex}{\mathbb{E}}

\renewcommand{\vec}[1]{\mathbf{#1}}
\newcommand{\etal}{\textit{et al}. }

\graphicspath{{images/}}
\title{\LARGE \bf Optimal Immunization Policy Using Dynamic Programming}

\graphicspath{{images/}}

\author{Atiye Alaeddini, and Daniel J. Klein
}

\begin{document}
	
	\maketitle
	\thispagestyle{empty}
	\pagestyle{empty}

	
	\begin{abstract}
		
Decisions in public health are almost always made in the context of uncertainty. Policy makers are responsible for making important decisions, faced with the daunting task of choosing from amongst many possible options.  This task is called planning under uncertainty, and is particularly acute when addressing complex systems, such as issues of global health and development. Uncertainty leads to cautious or incorrect decisions that cost time, money, and human life.  It is with this understanding that we pursue greater clarity on, and methods to address optimal policy making in health. Decision making under uncertainty is a challenging task, and all too often this uncertainty is averaged away to simplify results for policy makers. Our goal in this work is to implement dynamic programming which provides basis for compiling planning results into reactive strategies. We present here a description of an AI-based method and illustrate how this method can improve our ability to find an optimal vaccination strategy. We model the problem as a partially observable Markov decision process, POMDP and show how a re-active policy can be computed using dynamic programming. 
In this paper, we developed a framework for optimal health policy design in an uncertain dynamic setting. We apply a stochastic dynamic programing approach to identify the optimal time to change the health intervention policy and the value of decision relevant information for improving the impact of the policy.
		
		
	\end{abstract}

\section{Introduction}

The field of health economics is focused on maximizing the impact and cost effectiveness of health interventions.  These economic evaluations rely on mathematical or computational models to estimate the impact of candidate intervention packages, and associated incremental costs. Cost effectiveness analysis in health is an economic method that compares the lifetime costs and benefits associated with different health interventions. The optimal allocation of resources across health interventions is determined by solving a constrained optimization problem with the objective of maximizing health benefits subject to a budget constraint \cite{stinnett1996mathematical}. The health intervention we consider in this paper is vaccination, which is one of the primary intervention strategies used by public health agents to control infectious diseases.

Most of the current deterministic approaches used in vaccination planning focus on pre-determining vaccination strategies to reduce the expected number of infected population for a given budget. These types of policies are called open loop control. Another approach in optimal control is closed loop control or feedback control. Optimal planning using feedback control considers current data and find the optimal action for the most recent observed state. One advantage of feedback control is that corrections to process disturbances are automated. In this work, we extend the literature on optimal vaccination policies by developing an approach to optimal vaccination policy that is reactive to the data we receive over time. 

On the other hand, because funds are limited and data is expensive to collect, identifying the optimal information acquisition policy is a very important task. Good information is of significant value to health policy making programs, and we must weigh the potential value of information against other opportunity costs such as treatment, prevention, and system strengthening. Thus, balancing investments in information and investments on interventions will be important. The big question, here, is when more valuable to invest on gathering data rather than investing on intervention. When a lot of money, people, and process change would be involved, then careful methodical value of information modeling is likely to be necessary.

The first goal of this work is to find the optimal time for vaccination campaign in a region. Then the second step is to investigate when there is a value in collecting more information. This analysis helps to more efficiently determine what fraction of the funds of a disease control program should be allocated to conducting disease surveillance or collecting accurate data versus directed to increasing efforts to control the disease. Value of information and planning under uncertainty have a long history in control theory and computer science (AI) \cite{russell2002artificial,wiering2012reinforcement,sutton2018reinforcement}, but these methods have been rarely used in health policy making. To expand the utility and power of decision making under uncertainty, we propose to bring in ideas from artificial intelligence and optimal control theory. To demonstrate the effectiveness of the method, we will apply optimal planning algorithm on the problem of optimal SIA timing for Measles.


\section{Literature Review}

One of the fundamental problems in control theory is the linear quadratic Gaussian (LQG) control problem, which concerns linear systems driven by additive Gaussian noise. Output measurements are assumed to be corrupted by Gaussian noise. The LQG problem is to determine a feedback law that is optimal in the sense of minimizing the expected value of a quadratic cost criterion. To solve complex optimal control problems that do not fit in the well-developed LQG framework, we need a numerical method which falls in the dynamic programming category \cite{bertsekas1995dynamic}. Pioneered by Richard Bellman in the 1950s \cite{bellman1966dynamic}, dynamic programming enables complicated decision problems to be solved recursively, backwards in time, by breaking the problem into a series of smaller sub-problems. Although dynamic programming method involves discretization of the state and control spaces, but currently it is the best method that takes into account the stochastic dynamics and find global optimum. There are some local optimization methods, such as differential dynamic programming (DDP) \cite{jacobson1970differential} and iterative Linear Quadratic Regulator (iLQR) \cite{li2004iterative}, that don\rq{}t require state discretization. But the shortcoming of those methods is that they cannot deal with control constraints and non-quadratic cost functions and they find locally optimal control laws.
 
Our specific motivation of this work is finding the optimal health policy. The characteristics of problems in health are: high-dimensional nonlinear dynamics; control constraints (e.g. limited vaccination coverage); multiplicative noise; complex performance criteria that are rarely quadratic in the state variables. Due to all these complications, to date, the optimization in health policies have focused around finding fixed or time varying \cite{tanner2008finding,shattock2016interests} strategies which are pre-fixed for a specific period of time in the future. This work most closely align with cost optimization of immunization polices. In the works have been done in this field, different policies are evaluated by running many simulations and take the average of large number of simulations \cite{babigumira2011assessing,zimmermann2019optimization,bae2013economic}. One advantage of Dynamic programming is that it takes into account the uncertainty. Dynamic programming helps to investigate the effect of uncertainty on the optimal policy and determine the value of information in making more impactful decisions.

In classic control, the optimal control and optimal sensory system problems for linear systems are two separate problems, which are solved separately. However, we know that for nonlinear systems the observation policy changes the optimal control policy \cite{alaeddini2019augmented}. Finding optimal control and observation policy together for stochastic nonlinear system is a challenging problem. There exists works in literature studying optimal controlled observation for stochastic linear systems with quadratic running penalty \cite{wu2005optimal}. A computationally tractable algorithm has proposed to solve the stochastic sensor scheduling problem for the finite horizon linear quadratic Gaussian problem \cite{gupta2006stochastic}. Some works consider optimal sensor query by minimizing the uncertainty of state estimates \cite{krishnamurthy2007structured,bush2008computing,huang2019continuous}. In these works, the cost function comprises of estimation error and measurement cost.


In conventional system identification, the parameters of the system are estimated via well known Kalman filter techniques (EKF or UKF), usually by applying random controls, and after that the optimal control is computed via optimal control techniques for the estimated model. However, most of the time in public health there is high uncertainty in the disease model due to limited data. In order to remain robust to modeling errors, it would be better if we simultaneously perform estimation and control tasks. A Bayes-adaptive MDP (BAMDP) is a method used for planning under uncertain model parameters. BAMDP is basically a POMDP that the state has augmented with the unknown parameters defining the transition probabilities \cite{guez2013scalable,slade2017simultaneous}. In this work, we use a new method that formulates the parameter estimation problem as part of POMDP problem. The technique we used in this paper is very close to \cite{webb2014online} which plans control policies under model uncertainties.

\section{Methods}

\subsection{Optimal planning under uncertainty}

A classic planning problem in AI is specified as follows: Given a description of the current state of some system, a set of actions that can be performed on the system and a description of cost (or reward) of states and actions for the system, find a sequence of actions that can be performed to minimize the cumulative cost (or maximize the cumulative reward). The optimal planner must balance the potential of some plan achieving our goal to maximize the benefits against the cost of performing the plan.

Markov decision processes (MDPs) is an approach to planning under uncertainty
%
A policy, $\pi(s):s\rightarrow a$, for an MDP is a mapping from state to action that selects an action for each state. Now given a policy, we can define its value function, $V(s)$, which is equal to the immediate cost for the action selected by the policy at the current state plus the value function at the next state. 
A solution to an MDP is a policy that minimizes a term called the Hamiltonian, which is the MDP expected cost. 
%
Two popular methods for solving Hamiltonian and finding an optimal policy for an MDP are value iteration and policy iteration. In policy iteration, the current policy is repeatedly improved by finding some action in each state that has a higher value than the action chosen by the current policy for that state. The policy is initially chosen at random, and the process terminates when no improvement can be found \cite{puterman2014markov}. In value iteration, optimal policies are produced for successively longer finite horizons, until they converge, terminating when the maximum change in values between the current and previous value functions is below some threshold \cite{puterman2014markov}. 

Both of these methods, mentioned above, solve the MDP problem by recursively computing the optimal value function in a search tree containing approximately $|\mathcal{A}|^{K}$ possible sequences of moves, where $|\mathcal{A}|$ is the number of legal actions per state, and $K$ is the planning horizon. For large population, exhaustive search is hardly feasible. David Silver \etal introduced  new search algorithm that successfully combines neural network evaluations with Monte Carlo rollouts \cite{silver2016mastering,silver2017masteringChess,silver2017masteringGo}.

\subsection{Optimal planning using Hidden Markov model}

The MDP framework assumes that after each stochastic state transition, the state can be measured perfectly. In order to find the optimal vaccination policy using MDP, we need to perfectly observe the number of susceptible, infected, and recovered population at all time steps. But the information we receive is not perfect in practice. Depending on surveys, we only have access to part of states, e.g. we can observe a noisy measurement of the prevalence. In this work, we assume that the observation comes from a test with binary test characteristics $q=(q_1, q_2)$, where $q_1$ is the sensitivity, $q_2$ is the specificity, and $q_1+q_2>1$ . For a given survey coverage, $c_t$, we test $n_t = c_t \cdot N$ people for a disease at time $t$. The test could be imperfect. For a given sensitivity and test specificity, the observation can be modeled as:
\begin{equation} 
\begin{aligned}
	o_t|I_t &\sim \text{Binom} \left(n_t,\frac{q_1 I_t + (1-q_2)(S_t+R_t)}{N} \right)\,.
\end{aligned} \label{binom_obs}
\end{equation} 

In a Markov decision process the environment\rq{}s dynamics are fully determined by its current state, $s$. For any state and for any action, the transition probability determines the next state distribution, and the reward function determines the expected reward. In a partially observed Markov decision process (POMDP), the state cannot be directly observed by the agent. Instead, the agent receives an observation, determined by the observation probabilities, and the policy maps a history to a probability distribution over actions, where, history is a sequence of actions and observations. 

In POMDP we apply the very same idea as in MDP, but since the full states, $s$, are not observable, then the agent needs to choose the optimal policy only considering the belief state, $b$. The optimal policy at time t defined as:
\begin{equation}
\begin{aligned}
	& \pi_t^* \triangleq \underset{a \in \mathcal{A}}{\text{argmin}} \left[ l(b,a) + \alpha \int V_{t+1}(b^\prime) \Pr(b^\prime \mid b,a,o) d b^\prime \right]\,, \\
	& V_t (b) = \underset{a \in \mathcal{A}}{\text{min}} \left[ l(b,a) +  \alpha \int V_{t+1}(b^\prime) \Pr(b^\prime\mid b,a,o) d b^\prime \right]\,.
\end{aligned}
\end{equation}
Given a discount factor $\alpha \in (0\,, 1]$, the policy maker\rq{}s objective is to minimize the net cost of the policy over a given horizon (in case of finite horizon), and the given initial belief $b_0(x)$, and admissible possible decisions $a\in \mathcal{A}$. To compute the value function, $V(b)$, we need to probability of the next belief given current belief, $b$, if we take action $a$ and receive observation $o$. To compute this probability, we can use Bayes filter to update the belief. We need to compute two conditional probabilities; transition probability, $ \Pr(s^\prime \mid s,a)$, and conditional observation $\Pr(o\mid s)$. The transition probability can be calculated given the stochastic model of the disease, and the conditional observation can be computed using the formulation in \cref{binom_obs}.

The POMDP algorithms compute these value functions recursively. Finding optimal solution for POMDP is challenging. Some methods, like Monte Carlo POMDP \cite{thrun2000monte}, can approximate the value function of POMDP, and the optimal action can be read from the value function for any belief state. But the time complexity of solving POMDP value iteration is exponential in number of possible actions and observations, and the dimensionality of the belief space grows with number of states.

\subsection{Optimal vaccination and surveillance policies} \label{section:optimal_survey}

When a policy maker is attempting to solve a planning problem, the survey data are used to obtain noisy information. When faced with decisions in the presence of uncertainties, the policy maker should select the option with highest expected utility. By modeling this problem at a high level as a POMDP, the policy maker is able to account for the inherent uncertainty in the measurements. However, policy making in an uncertain environment often involves a trade off between exploratory actions, whose goal is to gather data, and regular actions which exploit the information gathered so far and pursue the objectives. In this situation the question is if we need more surveillance effort or increasing efforts on controlling disease (intervention vs. surveillance). 

The main assumption in both MDP and POMDP frameworks is that the agent continuously observes all states. This assumption does not capture many applications where observations are either limited or very costly. In many cases, we do not access data on a regular basis. For instance, in health, we only access to state information when there is a survey, and the uncertainty of the data also can vary depends on the testing sample size. A critical questions, here, are when is the best time to receive data and what sample size is required for a survey? There is no definitive answer to this question: large samples with rigorous selection are more powerful as they will yield more accurate results, but data collection and analysis will be proportionately more expensive. In this section, we consider the problem of optimal sample size and optimal time for survey.

Given a stochastic control problem for a POMDP in which there are a number of observation options available to us, with varying associated costs. The observation cost is added to the running cost and the optimal policy which is a combination of the optimal control and optimal sensor query is obtained from minimizing the total cost. The performance of the system depends on the level of uncertainty presented in the states estimation (belief). Thus, the controller needs to balance between performance and the penalty of requesting information. 


In the previous section, we assumed that the survey coverage is given. In this section, we define augmented control which includes control on observation. Thus, the control action is given by $a = \left( a_s \,, a_o \right)$, where $a_s $ is the vaccination fraction, similar to what we had in regular POMDP, and $a_o$ is the observation control action which determines the survey coverage. In this framework, the decision maker cannot see the observation continuously and has to determine the observation time (if necessary) and also the optimal vaccination strategy. There is trade off between observation and vaccination action. On one hand, observation facilitate more the decision maker to decide more efficiently that increases the system performance. But, on the other hand, higher sample size surveys cost more. 

At the beginning of each period $t$, the algorithm determines whether to invest on an intervention at time step $t$ and whether to spend the budget to conduct a survey over the period to obtain a better estimate of the belief $b$. Information, if sought is used together with the known stochastic dynamics of the disease to update the belief. Let $a_s(t) \in \{ 0, 1, 2, \cdots, A_s\}$ denote the intervention decision at time $t$, where $a_s=0$ indicates \emph{No Intervention} and $a_s=i \neq 0$ indicates \emph{Intervention} with $i$th level of vaccination coverage. The cost associated to this decision is monotonically increasing with the level of vaccination coverage, and is linear with respect to the number of vaccines required for the intervention. Here, we assumed that the vaccination cost is $c_v \cdot a_s$.

The amount of information collected is measured in terms of the survey coverage. The survey coverage determines the number of people being tested $a_o \in \{ 0, 1, 2, \cdots, A_o\}$; it is obtained at the cost $c_o \cdot a_o$ where $c_o$ is the cost of testing a person for the disease. Thus, at each time $t$ the policy maker implements the control $a = \left( a_s \,, a_o \right)$. The immediate cost for the current combination of state and control $l(s,a)$ is 
$$ l(s,a) = c_i (\bar{I}_t) + c_v a_o + c_o a_o \,,$$
where, $\bar{I}_t = I_t-I_{t-1}$ is the number of new cases from the previous time step $t-1$ to the current time $t$. To solve this augmented control POMDP, we need to define an augmented state:
$$ \bar{s} = (s, s_o)\,, $$
where $s_o = a_o$ is the augmented state.

Solving POMDP with the augmented control action, we have the optimal vaccination policy and optimal survey coverage at each time step. To simplify the problem, we can solve the POMDP for $a_s$ and $a_o$ are binary, which determines if we need to gather data (with a fixed survey coverage) or we need to have a vaccination campaign with a given fixed coverage.

\subsection{Estimate model parameters} \label{section:par_est}

The disease model often contain parameters, such as infectious rate, recovery rate, reporting rate, for which values are unknown. The value of these parameters are estimated from the observed data. The goal of this part is to estimate the parameters of the system, while still achieving other objectives. A benefit of this approach is that the costly re-calibration are to be avoided. Here, we used a method that formulates the parameter estimation problem as part of our POMDP problem, which plans policy that automatically trade off the effort spending on learning parameters and efforts spent achieving regular objectives. 

The disease models, e.g. SIR, SIS, or TSIR, have parameters which are either unknown or roughly estimated with limited data. Assume the model have a set of unknown parameters, called $\vec{p}$. We need to define an augmented state by appending the parameter-state, $\vec{p}$, to state space vector:
$$ \bar{s} = (s, s_\vec{p}) $$
where $s_\vec{p}$ is the parameters which are modeled as constant with stochastic white noise:
$$ \vec{p}_{t+1} = \vec{p}_t + \theta_t, \ \ \theta_t \sim \mathcal{N}(0, \Theta) \,.$$

The benefit of this approach is that we can estimate the parameters of the system and achieve our regular objectives together. One caveat is that adding parameters to the states space increases the size of the search space and we need to solve a larger POMDP compared to the regular POMDP.


\subsection{Linear Approximation of Value Function}

Solving a POMDP is finding a policy that maps each belief state into an action so that the expected sum of the discounted cost is minimized. In dynamic programming, the value function is parametrized by a finite set of hyperplanes over the belief space, which partition the belief space in a finite number of regions \cite{kaelbling1998planning}. These hyperplanes are called $\gamma$-vectors. Each of these $\gamma$-vectors maximizes the value function in a certain region and has an action associated with it. In this section we provide an approach that uses a form of dynamic programming in which a piecewise linear value function is approximated to predict the pruned set of alpha vectors.

Consider the problem of updating belief state, assuming we start at a particular belief state $b$. If we take action $a_i$ and receive observation $o_j$, then we can compute the next belief state,  $b^\prime$:
$$ b^\prime \sim \Pr(b^\prime \mid b,a_i,o_j) \,.$$
$$ b^\prime = \frac{\Pr(o \mid s^\prime) \sum_{s \in \mathcal{S}} \Pr(s^\prime \mid s, a) b(s)}{\Pr(o \mid b, a)} \,.$$
Since we assume that there are a finite number of actions and observations; given a belief state, there are a finite number of possible next belief states, each corresponding to a combination of action and observation. Given that our observations are probabilistic, each belief state has a probability associated with it. Before we take an action, each resulting belief state has a probability associated with it, and there are multiple possible next belief states (the number of observations for a given action). 
The value function for horizon $k \in \{1,\cdots,K\}$ is defined by:
\begin{equation}
\begin{aligned}
    &V_k(b) = \underset{a \in \mathcal{A}}{\text{min}} \left[ \Ex_o \left[V_{k+1}^{a,o}(b) \right] \right] \\
    &= \underset{a \in \mathcal{A}}{\text{min}} \left[ \sum_{s \in \mathcal{S}} l(s,a) b(s)+ \alpha \sum_{o \in \mathcal{O}}  \Pr(o \mid b,a) V_{k+1}(b^\prime) \right] \,,
\end{aligned}
\end{equation}
where,
$$ V_{k+1}^{a,o}(b) = \sum_{s \in \mathcal{S}} l(s,a) b+ \alpha V_{k+1}(b^\prime) \,.$$
If the final value of states at final horizon $K$ is given, then the value function for the last horizon $K$ can be written as:
$$ V_K(b) = \sum_{s \in \mathcal{S}} V_K(s) b(s) \,,$$
which is a linear function of $b$. Moving backwards in time, the value function for horizon $K-1$ is given by
\begin{equation} \label{linear_V}
 V_{K}^{a,o}(b) = \sum_{s \in \mathcal{S}} l(s,a) b+ \alpha V_K(s) b^\prime (s) \,.
\end{equation}
For a given action $a$ and observation $o$, the belief at step $K$ is a linear function of the belief in step $K-1$. If we can compute $V_{K}^{a,o}(b)$ for all combinations of $a \in \mathcal{A}$ and $o \in \mathcal{O}$, we can compute the value function at step $K-1$, $V_{K-1}(b)$. In words, \cref{linear_V} says that the value function for horizon $K-1$ is a combination of $|\mathcal{A}| \times |\mathcal{O}|$ linear functions. Following the same approach, we can show that the value function for each horizon can be obtained by minimizing a finite set of alpha vectors.
This means that, for each horizon $k$, the value function is obtained my computing the maximum of $|\mathcal{A}| \times |\mathcal{O}|$ linear functions. 

For the infinite horizon problem, the policy is stationary and the value function ($\gamma$-vactors are not depend on the time). Thus, we need to learn $|\mathcal{A}| \times |\mathcal{O}|$ linear functions to find the optimal policy. But, for a finite horizon problem, the policy is not stationary and varies with time (to final time). Therefore, to have an accurate value function for a finite horizon problem, we need to learn  $|\mathcal{A}| \times |\mathcal{O}| \times K$ linear functions, which can be very large, depends on the size of state space, action space, and the planning horizon. In this phase of the project, I compute  $|\mathcal{A}|$ per horizon. For a given belief at horizon $k$, the optimal control is obtained from the dot product of each $\gamma$-vector for that horizon. The optimal control is the minimum value from these dot products. 


\section{Optimal SIA timing for measles}

To achieve elimination and eradication goals against vaccine-preventable diseases, like measles, national routine immunization programs should be combined with intensive targeted Supplementary Immunization Activities (SIAs). 

To demonstrate the effectiveness of the AI-based approach for planning under uncertainty in this project, in this section we consider the problem of finding optimal time for supplemental immunization activities (SIA) for measles in Pakistan. 

The disease model we used in this work is the same as the model and all the assumptions presented in \cite{thakkar2019decreasing}, where, the measles in Pakistan is studied, and the optimal time for SIA in 2018 has been computed using the data on the number of cases in 2012 to 2018. They have shown that hundreds of thousands infectious can be averted only by changing the SIA time with no changes in campaign cost. To model the dynamics of measles, the TSIR (Time-series Susceptible Infected Recovered) model \cite{bjornstad2002dynamics} has been used. The TSIR model tracks the people in one of two categories: Susceptible ($S$) and Infectious ($I$). The stochastic discrete time formulation for TSIR epidemic model is given by:
\begin{equation*}
\begin{aligned}
	S_{t} &= (1-\mu_{t-1}) \left( B_{t} + S_{t-1} - I_{t} \right) \,, \\
	I_{t} &= \beta_{t} I_{t-1}^\alpha S_{t-1} \epsilon_t \,.
\end{aligned}
\end{equation*}
In TSIR setup, presented in \cite{thakkar2019decreasing}, the time step is 2 weeks. More details on the parameters of this disease model and how we can use data to calibrate the parameters of the model can be found in \cite{thakkar2019decreasing}. 

The state of the Markov process is defined as $s_t=(S_t, I_t, \tau)$, where $\tau = t \% 24$ is the time in a year. The action (decision) is the fraction of the susceptible population reached by SIA at time $t$. The goal, here, is to find the optimal time for SIA campaign. A simple form of the optimization problem can be written as:
\begin{equation*}
\begin{aligned}
& \underset{\mu_1, \mu_2, \cdots, \mu_K}{\text{minimize}}
& & \sum_{t=1}^K \alpha^t I_t \\
& \text{subject to}
& & \sum_{t=1}^K \mu_t \leq \text{budget} \,.
\end{aligned}
\end{equation*}
This optimization can be solved for:
\begin{itemize}
\item \emph{Infinite Horizon ($K \rightarrow \infty$)} In this case the discount factor should be smaller than one ($0<\alpha < 1$). The agent always has a constant expected amount of time remaining, so there is no reason to change action policy, and the optimal policy is stationary.
\item \emph{Finite Horizon (finite $K$)} In this case the discount factor can be equal to one ($0<\alpha \leq 1$). The optimal policy for finite horizon case is non-stationary: the way an agent chooses its actions on the last step is going to be very different from the way it chooses them when it has a long life ahead of it.
\item \emph{First Exit} the policy terminate as soon as we reach to a specific goal. 
\end{itemize}
Here we consider the finite horizon problem with discount factor equal to one. The challenge of finite horizon problem is that the optimal policy is not fixed, but it is a function of the time left until the end of the planning horizon. This requires that the policy need to be trained for every horizon. Although the dynamic programing solution for finite horizon is more difficult than infinite horizon, but, dynamic programming could save lots of time and energy to find the optimal plan. 

\section{Future Work}

This document proposed a technique for informative policy making in health. The developed method in this work addresses practical policy and program problems encountered by funders, governments, health planners, and program implementers to help them allocating limited resources more efficiently. The technique presented in this paper can be applied in problems regarding making decision under uncertainty. An example of this situation, which is the optimal measles SIA timing, was described in this paper. The application is not limited to this example. This technique can also be used in settings in which the decision maker wishes to identify the optimal time to stop the current intervention and initiate a new intervention, or evaluate the cost effectiveness of a new technology which improves the quality of data or efficiency of the intervention with some added cost. 

Many different adaptations, tests, and experiments have been left for the future. Future work concerns deeper analysis of particular mechanisms and new proposals to try different methods.
There are some ideas that I would like to continue working on, such as controlled observation \cref{section:optimal_survey} and simultaneous parameter estimation \cref{section:par_est}. In addition, the following ideas could be tested:
\begin{itemize}
\item \emph{implementing Monte Carlo based methods to sample from tree search} \\
Solving with dynamic programing requires discretization of the states. If we add both the controlled observation and the model parameters to the state, then the augmented state become very high dimensional, and the transition and observation matrices are very large. We can use Monte Carlo sampling to break the curse of dimensionality for belief state updates. This method is called POMCP and has first introduced in \cite{silver2010monte}. POMCP is an extension of the Monte Carlo Tree Search algorithm that was previously implemented for large MDPs in AlphaGo.

\item \emph{training transition and observation matrices directly from data, instead of using disease model}  \\
The TSIR model has been used in dynamic programming to generate the transition and observation matrices. The TSIR model estimates the model parameters from a linear regression, given the time series of data. Basically, we do not need the disease model in dynamic programming. We can directly use the data to estimate the state transition and conditional observation. Given an observation sequence (time series of the number of cases), we can train a discrete Hidden Markov Model and find the initial belief, conditional observation and state transition matrices using maximum likelihood algorithm \cite{rabiner1989tutorial}.

\end{itemize}




\begin{appendices}

\end{appendices}

\bibliographystyle{ieeetr}
\bibliography{citations}

\end{document}